\documentclass[letterpaper]{article}
\usepackage{times}
\usepackage[stable]{footmisc}
\usepackage{graphicx}
\usepackage{breakcites}
\usepackage{algorithm,algorithmic,a4wide,amssymb,multicol,enumitem,url}
\usepackage{amsmath}
\usepackage[breaklinks]{hyperref}
 \usepackage[top=3.7cm, bottom=3.7cm, left=3.7cm, right=3.7cm]{geometry} 
\makeatletter
\renewcommand\hyper@natlinkbreak[2]{#1}
\makeatother
\newcommand{\mytilde}{\raise.17ex\hbox{$\scriptstyle\mathtt{\sim}$}}
\begin{document}

\title{On Learning to Think: Algorithmic Information Theory for
Novel Combinations of Reinforcement Learning Controllers and Recurrent Neural World Models \\
{\small Technical Report}
}

\date{30 November 2015}
\author{J\"{u}rgen Schmidhuber~\\
The Swiss AI Lab  \\
Istituto Dalle Molle di Studi sull'Intelligenza  Artificiale (IDSIA)\\
Universit\`{a} della Svizzera italiana (USI) \\
Scuola universitaria professionale della Svizzera italiana (SUPSI) \\
Galleria 2, 6928 Manno-Lugano, Switzerland \\ 
}
\maketitle

\begin{abstract}
  This paper addresses the general problem of reinforcement learning
  (RL) in partially observable environments.  In 2013,
  our large RL recurrent neural networks (RNNs) learned from scratch
  to drive simulated cars from high-dimensional video input. However,
  real brains are  more powerful
  in many ways. In particular, they learn a predictive model of their
  initially unknown environment, and somehow use it for abstract (e.g., hierarchical)
  planning and reasoning.  Guided by algorithmic information theory,
  we describe RNN-based AIs (RNNAIs) designed to do the same. Such an
  RNNAI can be trained on never-ending sequences of tasks, some of
  them provided by the user, others invented by the RNNAI itself in a
  curious, playful fashion, to improve its RNN-based world model. 
  Unlike our previous model-building RNN-based RL machines dating back to 1990, the
  RNNAI learns to actively query its model for abstract reasoning and planning and decision
  making, essentially ``learning to think."
The basic ideas of this report 
can be applied to many other cases where 
one RNN-like system exploits the algorithmic information content of another. They
are taken from a grant
  proposal submitted in Fall 2014, and
also explain concepts such as ``mirror neurons."   Experimental results
  will be described in separate papers.
\end{abstract}

\newpage
\tableofcontents

\section{Introduction to Reinforcement Learning (RL) with Recurrent\\
  Neural Networks (RNNs) in Partially Observable Environments\footnote{Parts
    of this introduction are similar to parts of a much more extensive recent
{\em Deep Learning} overview \cite{888} which has many additional references.}}
\label{intro}

General {\em Reinforcement Learning} (RL) agents must discover,
without the aid of a teacher, how to interact with a dynamic,
initially unknown, partially observable environment in order to
maximize their expected cumulative reward signals, e.g., \cite{Kaelbling:96,Sutton:98,wiering2012}.  There may be
arbitrary, {\em a priori} unknown delays between actions and perceivable
consequences.  The RL problem is as hard as any problem of computer
science, since any task with a computable description can be
formulated in the RL framework, e.g., \cite{Hutter:05book+}.

To become a general problem solver that is able to run arbitrary
problem-solving programs, the controller of a robot or an artificial
agent must be a general-purpose
computer~\cite{Goedel:31,Church:36,Turing:36,Post:36}.  Artificial
recurrent neural networks (RNNs) fit this bill.  A typical RNN
consists of many simple, connected processors called neurons, each
producing a sequence of real-valued activations.  Input neurons get
activated through sensors perceiving the environment, other neurons
get activated through weighted connections or wires from previously
active neurons, and some neurons may affect the environment by
triggering actions.  {\em Learning} or {\em credit assignment} is
about finding real-valued weights that make the NN exhibit {\em
  desired} behavior, such as driving a car.  Depending on the problem
and how the neurons are connected, such behavior may require long
causal chains of computational stages, where each stage transforms
the aggregate activation of the network, often in a non-linear manner.

Unlike feedforward NNs (FNNs;~\cite{Hertz:91,bishop:2006}) and {\em
  Support Vector Machines} (SVMs;~\cite{Vapnik:95,advkernel}), RNNs
can in principle interact with a dynamic partially observable environment in arbitrary, computable
ways, creating and processing memories of sequences of input
patterns~\cite{siegelmann91turing}.  The weight matrix of an RNN is
its program.  Without a teacher, reward-maximizing programs of an RNN
must be learned through repeated trial and error.

\subsection{RL through Direct and Indirect Search in RNN Program Space}
\label{direct}

It is possible to train small RNNs with a few 100 or 1000 weights
using evolutionary
algorithms~\cite{Rechenberg:71,Schwefel:74,Holland:75,Fogel:66,goldberg:gabook89}
to search the space of 
NN weights~\cite{miller:icga89,wieland1991,cliff1993,yao:review93,nolfi:alife4,Sims:1994:EVC,yamauchi94sequential,miglino95evolving,moriarty:phd,gomez:phd,Gomez:03,pasemann99,juang2004,wierstraCEC08,glasmachers:2010b,Sun2009a,sun:gecco13,whiteson2012},
or through policy gradients
(PGs)~\cite{Williams:86,Williams:88,Williams:92,Sutton:99,baxter2001,aberdeenthesis,ghavamzadehICML03,stoneICRA04,wierstraCEC08,rueckstiess2008b,peters2008neuralnetworks,peters2008neurocomputing,sehnke2009parameter,gruettner2010multi,wierstra2010,peters2010,grondman2012,heess2012}\cite[Sec.~6.6]{888}. For
example, our evolutionary algorithms outperformed traditional,
{\em Dynamic Programming}~\cite{Bellman:1957}-based RL methods
\cite{Sutton:98}\cite[Sec.~6.2]{888} in partially observable environments, e.g.,
\cite{Gomez:08jmlr}.  However, these techniques by themselves are
insufficient for solving complex control problems involving
high-dimensional sensory inputs such as video, from scratch.  The
program search space for networks of the size required for these tasks
is simply too large.

However, the search space can often be reduced dramatically by evolving
{\em compact encodings} of neural networks (NNs), e.g., through {\em
  Lindenmeyer Systems}~\cite{lindenmayer94}, {\em graph
  rewriting}~\cite{kitano90}, {\em Cellular
  Encoding}~\cite{gruau:tr96-048}, {\em HyperNEAT}~\cite{stanley09},
and other techniques~\cite[Sec.~6.7]{888}.  In very general early work,
we used {\em universal assembler-like languages} to encode
NNs~\cite{Schmidhuber:97nn+}, later {\em coefficients of a Discrete
  Cosine Transform (DCT)}~\cite{koutnik:gecco13}.  The latter method,
{\em Compressed RNN Search}~\cite{koutnik:gecco13}, was used to
successfully evolve RNN controllers with over a million weights (the
largest ever evolved) to drive a simulated car in a video game, based
solely on a high-dimensional video
stream~\cite{koutnik:gecco13}---learning both control and visual
processing from scratch, without unsupervised
pre-training of a vision system.  
This was the first published Deep
Learner to learn control policies directly from high-dimensional
sensory input using RL.

One can further facilitate the learning task of controllers through
certain types of supervised learning (SL) and unsupervised learning
(UL) based on gradient descent techniques.  In particular, UL/SL can
be used to compress the search space, and to build predictive world
models to accelerate RL, as will be discussed later.  But first let us
review the relevant NN algorithms for SL and UL.

\subsection{Deep Learning in NNs: Supervised \& Unsupervised Learning (SL \& UL)}
\label{SLUL}

The  term {\em Deep Learning} was first introduced 
to Machine Learning in 1986~\cite{dechter1986deeplearning}
and to NNs in 2000~\cite{aizenberg2000deeplearning,scholarpedia2015}.
The first deep learning NNs, however, date back to 
the 1960s~\cite{ivakhnenko1965,888}
(certain more recent developments are covered in a survey~\cite{lecun2015nature}).

To maximize differentiable objective functions of SL and UL, NN
researchers almost invariably use backpropagation
(BP)~\cite{Kelley:1960,bryson:1961,dreyfus:1962} in discrete graphs of
nodes with differentiable activation functions
~\cite{Linnainmaa:1970,SPEELPENNING80A}\cite[Sec.~5.5]{888}.  Typical
applications include BP in FNNs~\cite{Werbos:81sensitivity}, or BP
through time (BPTT) and similar methods in RNNs, e.g.,
\cite{Werbos:88gasmarket,WilliamsZipser:92,RobinsonFallside:87tr}\cite{888}.
BP and BPTT suffer from the {\em Fundamental Deep Learning Problem} 
first discovered and analyzed in my lab in 1991: with standard activation
functions, cumulative backpropagated error signals decay exponentially
in the number of layers, or they
explode~\cite{Hochreiter:91,Hochreiter:01book}.  Hence most early
FNNs~\cite{Werbos:81sensitivity,Rumelhart:86} had few layers.
Similarly, early RNNs~\cite[Sec.~5.6.1]{888} could not generalize well
under both short and long time lags between relevant events.  Over the
years, several ways of overcoming the {\em Fundamental Deep Learning Problem} have been explored.  For
example, deep stacks of unsupervised RNNs~\cite{chunker91and92} or
FNNs~\cite{ballard1987modular,HinSal06,lecun2015nature} help to
accelerate subsequent supervised learning through
BPTT~\cite{chunker91and92,schmidhuber1993} or
BP~\cite{HinSal06}.
One can also ``distill" or compress the knowledge of a teacher RNN  into a student RNN  by 
forcing the student to predict the hidden units of the teacher~\cite{chunker91and92,schmidhuber1993}. 

 {\em Long Short-Term Memory}
(LSTM;~\cite{lstm97and95,Gers:2000nc,Graves:09tpami})
alleviates the  {\em Fundamental Deep Learning Problem}, and was the first RNN architecture to
win international contests (in connected handwriting), e.g.,
\cite{graves:2009nips,schmidhuber2011agi}\cite{888}.   {\em
  Connectionist Temporal Classification} (CTC)~\cite{Graves:06icml}
is a widely used gradient-based method for finding RNN weights that
maximize the probability of teacher-provided label sequences, given
(typically much longer and more high-dimensional) streams of
real-valued input vectors.  For example, CTC was used by Baidu  to break an
important speech recognition record~\cite{hannun2014}.  Many recent
state-of-the-art results in sequence processing are based on LSTM,
which learned to control robots~\cite{mayer2008},
and was used to set benchmark records in  prosody contour
prediction~\cite{fernandez2014} (IBM), 
 text-to-speech synthesis~\cite{fan2014}
(Microsoft),  large
vocabulary speech recognition~\cite{sak2014large} (Google), and
machine translation~\cite{sutskever2014} (Google).  
CTC-trained LSTM greatly  improved Google Voice~\cite{sak2015voice} and is now available to over a billion smartphone users.
Nevertheless, at
least in some applications, other RNNs  may sometimes yield better
results than gradient-based
LSTM~\cite{Martens:2011hessfree,DBLP:conf/icann/SchaferUZ06,DBLP:series/lncs/ZimmermannTG12,Jaeger:04,Schmidhuber:07nc,pascanu2013,icml2014}.
Alternative NNs with differentiable memory have been proposed~\cite{Schmidhuber:92ncfastweights,Das:92,mozer1993connectionist,Schmidhuber:93selfreficann,Schmidhuber:93ratioicann,Hochreiter:01meta,graves2014ntm,weston2014memory}.

Today's faster computers, such as GPUs, mitigate the {\em Fundamental Deep Learning Problem} for  
FNNs~\cite{gpu2004,chellapilla:2006b,raina2009large,ciresan:2010,ciresan:2011ijcai}. In
particular, many recent computer vision contests were won by fully
supervised Max-Pooling Convolutional NNs (MPCNNs),
which consist of alternating
convolutional~\cite{Fukushima:1979neocognitron,Behnke:LNCS}
and max-pooling~\cite{weng1992} layers topped off by standard fully
connected output layers. All weights are trained by
backpropagation~\cite{LeCun:89,ranzato-cvpr-07,scherer:2010,888}. 
Ensembles~\cite{Schapire:90,breiman:1996}
of GPU-based MPCNNs~\cite{ciresan:2011ijcai,ciresan:2011ijcnn} 
achieved dramatic improvements of long-standing benchmark records,
e.g., MNIST (2011), won numerous competitions
~\cite{schmidhuber2011agi,ciresan:2010,ciresan:2011ijcnn,ciresan:2011icdar,ueli:2011icdar,ciresan2012cvpr,ciresan2012nips,Krizhevsky:2012,zeiler2013,miccai2013,888},
and achieved the first human-competitive or even superhuman results on
well-known benchmarks,
e.g.,~\cite{schmidhuber2011agi,ciresan2012cvpr,888}.  There are many
recent variations and
improvements~\cite{malik2013,goodfellow2014multi,karpathy2014,goodfellow2013maxout,szegedy2014,srivastava2013compete,888}.
Supervised {\em Transfer 
Learning} from one dataset to another~\cite{caruana1997,Ciresan:2012a} can speed up learning.
A combination of Convolutional NNs (CNNs) and LSTM led to best results in automatic
image caption generation~\cite{vinyals2014caption}.

\subsection{Gradient Descent-Based NNs for RL}
\label{GDRL}

Perhaps the most well-known RL application is Tesauro's backgammon
player~\cite{Tesauro:94} from 1994 which learned to achieve the level
of human world champions, by playing against itself.  It uses a
reactive (memory-free) policy based on the simplifying assumption of
{\em Markov Decision Processes}: the current input of the RL
agent conveys all information necessary to compute an optimal next
output event or decision. The policy is implemented as a
gradient-based FNN trained by the method of temporal
differences~\cite{Sutton:98}\cite[Sec.~6.2]{888}.  During
play, the FNN learns to map board states to predictions of expected
cumulative reward, and selects actions leading to states with maximal
predicted reward.  A very similar approach (also based on over
20-year-old methods) employed a CNN (see Sec.~\ref{SLUL}) to play
several Atari video games directly from 84$\times$84 pixel 60 Hz video
input~\cite{atari2015}, using {\em Neural Fitted Q-Learning} (NFQ)~\cite{nfq}
 based on experience replay (1991)~\cite{Lin:91}.  Even
better results were achieved by using (slow) Monte Carlo tree planning
to train comparatively fast deep NNs~\cite{atarimcts2014}.

Such FNN approaches cannot work in realistic partially observable environments where memories of
previous inputs have to be stored for {\it a priori} unknown time
intervals. This triggered work on {\em partially observable Markov decision
  problems}
(POMDPs)~\cite{Schmidhuber:90sandiego,Schmidhuber:90cmss,Schmidhuber:91nips,Ring:91,Ring:93,Ring:94,Williams:92,Lin:93,Teller:94,Kaelbling:95,Littman:95,Boutilier:96,Jaakkola:95,McCallum:96,kimura1997,Wiering:96levin,Wiering:97ab,otsuka2010}.
Traditional RL
techniques~\cite{Sutton:98}\cite[Sec.~6.2]{888} based on
{\em Dynamic Programming}~\cite{Bellman:1957} can be combined
with gradient descent methods to train an RNN as a value-function
approximator that maps entire event histories to predictions of
expected cumulative reward~\cite{Schmidhuber:91nips,Lin:93}.
LSTM~\cite{lstm97and95,Gers:2000nc,Perez:02,graves05nn,Graves:09tpami}
(see Sec.~\ref{SLUL}) was used in this way for RL
robots~\cite{Bakker:03robot}.

Gradient-based UL may be used to reduce an RL controller's search
space by feeding it only compact codes of high-dimensional
inputs~\cite{Jodogne07,Legenstein2010,cuccu2011}\cite[Sec.~6.4]{888}.
For example, NFQ~\cite{nfq} was applied to real-world control
tasks~\cite{lange,rieijcnn12} where purely visual inputs were
compactly encoded in hidden layers of deep autoencoders \cite[Sec.~5.7
and and 5.15]{888}. RL combined with unsupervised learning based on
{\em Slow Feature
  Analysis}~\cite{WisSej2002,DBLP:journals/neco/KompellaLS12} enabled
a humanoid robot to learn skills from raw video
streams~\cite{luciwkomp13}.  A RAAM RNN~\cite{pollack1988implications}
was employed as a deep unsupervised sequence encoder for
RL~\cite{Gisslen2011agi}.

\subsubsection{Early RNN Controllers with Predictive RNN World Models}
\label{earlyCM}

One important application of gradient-based UL is to obtain a
predictive world model, $M$, that a controller, $C$, may use to
achieve its goals more efficiently, e.g., through cheap, ``mental"
$M$-based trials, as opposed to expensive trials in the real
world~\cite{Werbos:89identification,Sutton:90dyna}.  The first
combination of an RL RNN $C$ and an UL RNN $M$ was ours and dates back to
1990~\cite{Schmidhuber:90sandiego,Schmidhuber:90cmss,Schmidhuber:90sab,Schmidhuber:91nips},
generalizing earlier similar controller/model systems ($CM$ systems)
based on FNNs~\cite{Werbos:87,NguyenWidrow:89}; compare related
work~\cite{Munro:87,Jordan:88,Werbos:89identification,Werbos:89neurocontrol,RobinsonFallside:89,JordanRumelhart:90,narendra1990,Werbos:92sticky,kawato1993,cochocki1993,levin1995,miller1995,ljung1998,prokhorov2001,ge2010}\cite[Sec.~6.1]{888}.
$M$ tries to learn to predict $C$'s inputs (including reward signals)
from previous inputs and actions.  $M$ is also temporarily used as a
surrogate for the environment: $M$ and $C$ form a coupled RNN where
$M$'s outputs become inputs of $C$, whose outputs (actions) in turn
become inputs of $M$.  Now a gradient descent
technique~\cite{Werbos:88gasmarket,WilliamsZipser:92,RobinsonFallside:87tr}(see
Sec.~\ref{SLUL}) can be used to learn and plan ahead by training
$C$ in a series of $M$-simulated trials to produce output action
sequences achieving {\em desired input events}, such as high
real-valued reward signals (while the weights of $M$ remain fixed).
An RL active vision system, from 1991~\cite{SchmidhuberHuber:91},
used this basic principle to learn sequential shifts (saccades) of a
fovea to detect targets in a visual scene, thus learning a rudimentary
version of selective attention.

Those early $CM$ systems, however, did not yet use powerful RNNs such as
LSTM.  A more fundamental problem is that if the environment is too
noisy, $M$ will usually only learn to approximate the conditional
expectations of predicted values, given parts of the history. In
certain noisy environments, Monte Carlo Tree Sampling
(MCTS;~\cite{browne2012}) and similar techniques may be applied to $M$
to plan successful future action sequences for $C$.  All such methods,
however, are about simulating possible futures time step by time step,
without profiting from human-like hierarchical planning or abstract
reasoning, which often ignores irrelevant details.

\subsubsection{Early Predictive RNN World Models Combined with Traditional RL}
\label{earlyMtradRL}

In the early 1990s, an RNN $M$ as in Sec.~\ref{earlyCM} was also
combined~\cite{Schmidhuber:91nips,Lin92memoryapproaches} with
traditional temporal difference methods
\cite{Kaelbling:95,Sutton:98}\cite[Sec.~6.2]{888} based on
the Markov assumption (Sec.~\ref{GDRL}).  While $M$ is processing the
history of actions and observations to predict future inputs and
rewards, the internal states of $M$ are used as inputs to a temporal
difference-based predictor of cumulative predicted reward, to be
maximized through appropriate action sequences.  One of our systems
described in 1991~\cite{Schmidhuber:91nips} actually collapsed the
cumulative reward predictor into the predictive world model, $M$.

\subsection{Hierarchical \& Multitask RL and Algorithmic Transfer Learning}\label{hie}

Work on NN-based {\em Hierarchical RL} (HRL) without predictive world
models has been published since the early 1990s. In particular,
gradient-based {\em subgoal discovery} with RNNs decomposes RL tasks
into subtasks for submodules~\cite{Schmidhuber:91icannsubgoals}.
Numerous alternative HRL techniques have been
proposed~\cite{Ring:91,Ring:94,Jameson:91,TenenbergKarlssonWhitehead,Weiss:94a,partigame,Precup:MTimeNIPS98,Dietterich:MAXQ,menache2002,DoyaSamejimaKatagiriKawato,barto2003hrl,SamejimaDoyaKawato,Bakker:04ias,simsek2008skill}.
While HRL frameworks such as {\em Feudal RL}~\cite{Dayan:93} and {\em
  options}~\cite{sutton1999between,Barto:04,Singh:05nips} do not
directly address the problem of automatic subgoal discovery, {\em
  HQ-Learning}~\cite{Wiering:97ab} automatically decomposes problems
in partially observable environments into sequences of simpler subtasks that can be solved by
memoryless policies learnable by reactive sub-agents.  Related methods
include incremental NN evolution~\cite{gomez:ab97},
hierarchical evolution of
NNs~\cite{stoneML05,vanhoorn:09cig}, and hierarchical Policy
Gradient algorithms~\cite{ghavamzadehICML03}.  Recent HRL
organizes potentially deep NN-based RL sub-modules into
self-organizing, 2-dimensional motor control
maps~\cite{ring:icdl2011} inspired by neurophysiological
findings~\cite{Graziano:book}.  The methods above, however, assign
credit in hierarchical fashion by limited fixed schemes that are not
themselves improved or adapted in problem-specific ways.  The next
sections will describe novel $CM$ systems that overcome such drawbacks of
above-mentioned methods.

General methods for incremental multitask RL and algorithmic transfer learning that are not NN-specific include 
the evolutionary ADATE system~\cite{Olsson:95},
the {\em Success-Story Algorithm for Self-Modifying Policies} running on general-purpose computers~\cite{Schmidhuber:94self,Schmidhuber:97bias,Schmidhuber:97ssa},
and 
the {\em Optimal Ordered Problem Solver}~\cite{Schmidhuber:04oops}, which learns
algorithmic solutions to new problems by inspecting and exploiting (in arbitrary computable fashion)
solutions to old problems,
in a way that is asymptotically time-optimal.
And  {\sc PowerPlay}~\cite{Schmidhuber:13powerplay,Srivastava2013first} 
incrementally learns to become a more and more general algorithmic problem solver, by
continually searching the space of possible pairs of new tasks
  and modifications of the current solver, until it finds a more
powerful solver that, unlike the unmodified solver, solves all
previously learned tasks plus the new one, or at least 
simplifies/compresses/speeds up previous solutions, without forgetting
any.

\section{Algorithmic Information Theory (AIT) for RNN-based AIs}
\label{AIT}

Our early RNN-based $CM$ systems (1990) mentioned in Sec.~\ref{earlyCM} learn a
predictive model of their initially unknown environment. 
Real brains seem to do so too, but are still far superior to present artificial systems in many ways.  
They seem to
exploit the model in smarter ways, e.g., to plan action sequences in
hierarchical fashion, or through other types of abstract reasoning,
continually building on earlier acquired skills, becoming increasingly
general problem solvers able to deal with a large number of diverse
and complex tasks.  Here we describe RNN-based Artificial Intelligences (RNNAIs)
designed to do the same by ``learning to think."\footnote{The terminology is partially inspired
by  our RNNAISSANCE workshop at NIPS 2003 ~\cite{Schmidhuber:03rnnaissance}.}

While FNNs are traditionally linked~\cite{bishop:2006} to concepts of
statistical mechanics and information theory
~\cite{boltzmann1909,Shannon:48,kullback1951}, the programs of general computers such as RNNs
call for the framework of Algorithmic Information Theory (AIT)
~\cite{Solomonoff:64,Kolmogorov:65,Chaitin:66,Levin:73a,Solomonoff:78,LiVitanyi:97} (own AIT
work:~\cite{Schmidhuber:95kol+,Schmidhuber:97nn+,Schmidhuber:02ijfcs,Schmidhuber:02colt,Schmidhuber:04oops}).
Given some universal programming language~\cite{Goedel:31,Church:36,Turing:36,Post:36} for a universal computer,
 the algorithmic information content or Kolmogorov complexity of some computable object is 
the length of the shortest program that computes it. 
Since any program for one computer can be translated into a functionally equivalent program for a different computer by a compiler program of constant size, 
the Kolmogorov complexity of most objects hardly depends on the particular computer used. 
Most computable objects of a given size, however, are hardly compressible, since there are only relatively few programs that are much shorter. 
Similar observations hold for practical variants of Kolmogorov complexity that explicitly take into account program runtime~\cite{Levin:73,Allender:92,Watanabe:92,LiVitanyi:97,Schmidhuber:97nn+,Schmidhuber:02colt}.
Our RNNAIs are inspired by the following argument.

\subsection{Basic AIT Argument}
\label{argument}

According to AIT, given some universal computer, $U$, whose programs are
encoded as bit strings, the mutual information between two programs
$p$ and $q$ is expressed as $K(q \mid p)$, 
the length of the shortest program
$\bar{w}$ that computes $q$, given $p$, ignoring an additive constant
of $O(1)$ depending on $U$ (in practical applications the computation
will be time-bounded~\cite{LiVitanyi:97}). That is, if $p$ is a
solution to problem $P$, and $q$ is a fast (say, linear time) solution
to problem $Q$, and if $K(q \mid p)$ is small, and $\bar{w}$ is both fast
and much shorter than $q$, then {\em asymptotically optimal universal
  search}~\cite{Levin:73,Schmidhuber:04oops} for a solution to $Q$,
given $p$, will generally find $\bar{w}$ first (to compute $q$ and
solve $Q$), and thus solve $Q$ much faster than search for $q$ from
scratch~\cite{Schmidhuber:04oops}.

\subsection{One RNN-Like System Actively Learns to Exploit Algorithmic Information of Another}  
\label{general}

The AIT argument \ref{argument} above has broad applicability. 
Let both $C$ and $M$ be RNNs or similar general parallel-sequential  computers~\cite{Schmidhuber:92ncfastweights,Das:92,mozer1993connectionist,Schmidhuber:93selfreficann,Schmidhuber:93ratioicann,Hochreiter:01meta,graves2014ntm,weston2014memory}.
$M$'s vector of learnable real-valued parameters $w_M$ is trained by any SL or UL or RL algorithm to perform a certain well-defined task in some environment. Then  $w_M$ is frozen.
Now the goal is to train $C$'s parameters  $w_C$ by some learning algorithm to perform another well-defined task
whose solution may share mutual algorithmic information with the solution to $M$'s task. 
To facilitate this,  we simply allow $C$ to learn to actively inspect and reuse (in essentially arbitrary computable fashion)
the algorithmic information conveyed by $M$ and $w_M$. 

Let us consider a trial during which $C$ makes an attempt to solve its 
given task within a series of  discrete time steps $t = t_a, t_a+1, \ldots, t_b$.  
$C$'s learning algorithm may use the experience gathered during the trial
to modify  $w_C$ in order to improve $C$'s performance in later trials.
During the trial, we give $C$ an opportunity to explore and exploit or ignore $M$ by interacting with  it. 
In what follows, $C(t)$, $M(t)$, $sense(t)$, $act(t)$, $query(t)$, $answer(t)$, $w_M$,  $w_C$ denote vectors of real values;
$f_C, f_M$ denote
computable~\cite{Goedel:31,Church:36,Turing:36,Post:36} functions. 

At any time $t$, 
$C(t)$ and  $M(t)$ denote 
$C$'s and $M$'s current  {\em states}, respectively. 
They may represent current neural activations or fast weights~\cite{Schmidhuber:92ncfastweights,Schmidhuber:93selfreficann,Schmidhuber:93ratioicann}
 or other dynamic variables that may change during information processing.
 $sense(t)$ is the current input  from the environment (including reward signals if any);
a part of $C(t)$ encodes 
the current output $act(t)$ to the environment, another a memory of previous events (if any).
Parts of $C(t)$ and $M(t)$ intersect in the sense that both $C(t)$ and $M(t)$ also encode
$C$'s current $query(t)$ to $M$,
and $M$'s current $answer(t)$ to $C$ (in response to previous queries),
thus  representing an  interface between $C$ and $M$.

$M(t_a)$ and $C(t_a)$ are initialized by default values.
For $t_a \leq t< t_b$, 
\[
C(t+1) = f_C(w_C,  C(t), M(t), sense(t), w_M)
\]
with learnable parameters $w_C$;
$act(t)$  is a computable function of $C(t)$ and may influence  $in(t+1)$,
and $M(t+1)=f_M(C(t), M(t), w_M)$ with fixed parameters $w_M$.
So both $M(t+1)$ and $C(t+1)$ are computable functions of  previous events including queries and answers
transmitted through the learnable $f_C$.

According to the AIT argument, 
provided that $M$ conveys substantial algorithmic information about $C$'s task,
and the trainable interface $f_C$ between $C$ and $M$ allows $C$ to address and extract and exploit this information quickly,
and $w_C$ is small compared to the fixed $w_M$,
the search space of 
$C$'s learning algorithm (trying to find a good  $w_C$ through a series of trials) 
should be much smaller than the one of a similar competing system $C'$ 
that has no opportunity to  query $M$ but has to learn the task from scratch.

For example, suppose that $M$ has learned to represent (e.g., through predictive coding~\cite{chunker91and92,SchmidhuberHeil:96})
 videos of people placing toys in boxes,
or to summarize such videos through textual outputs.
Now suppose $C$'s task is to learn to control a robot that places toys in boxes. 
Although the robot's actuators may be quite different from human arms and hands, 
and although videos and video-describing texts are quite different from desirable trajectories of
robot movements, $M$ is expected to convey algorithmic information about $C$'s task, perhaps in form of connected
high-level spatio-temporal  feature detectors representing typical movements of hands and elbows independent of  arm size.  
Learning a $w_C$ that addresses and extracts this information from $M$ and partially reuses it to solve the robot's task may 
be much faster than learning to solve the task from scratch without access to $M$. 

The setups of Sec.~\ref{think} are  special cases of the general scheme in the present Sec. \ref{general}.

\subsection{Consequences of the AIT Argument for Model-Building Controllers}
\label{outline}

The simple AIT insight above suggests that in many partially observable environments it should be
possible to greatly speed up the program search of an RL RNN, $C$, by
letting it learn to access, query, and exploit in arbitrary
computable ways the program of a typically much bigger gradient-based
UL RNN, $M$, used to model and compress the RL agent's entire growing
interaction history of all failed and successful trials.

Note that the $\bar{w}$ of  Sec.~\ref{argument} may implement
all kinds of well-known, computable types of reasoning, e.g., by 
hierarchical reuse of subprograms of $p$~\cite{Schmidhuber:04oops}, by
analogy, etc. That is, we may perhaps even expect $C$ to learn
to exploit $M$ for human-like abstract thought.

Such novel $CM$ systems will be a central topic of Sec.~\ref{CM}.
Sec.~\ref{curiosity} will also discuss exploration based on
efficiently improving $M$ through $C$-generated experiments.

\section{The RNNAI and its Holy Data}
\label{holy}

In what follows, let $m,n,o$ denote positive integer constants, and
$i,k,h,t,\tau$ positive integer variables assuming ranges implicit
in the given contexts.  The $i$-th component of any real-valued vector,
$v$, is denoted by $v_i$.  Let the RNNAI's life span a discrete sequence of
time steps, $t = 1,2,\ldots, t_{death}$.

At the beginning of a given time step, $t$, there is a ``normal'' sensory
input vector, $in(t) \in \mathbb{R}^m$, and a reward input vector, $r(t)
\in \mathbb{R}^n$.  For example, parts of $in(t)$ may represent the pixel
intensities of an incoming video frame, while components of $r(t)$ may
reflect external positive rewards, or negative values produced by pain
sensors whenever they measure excessive temperature or pressure.  Let
$sense(t) \in \mathbb{R}^{m+n}$ denote the concatenation of the
vectors $in(t)$ and $r(t)$.  The total reward at time $t$ is $R(t)=
\sum_{i=1}^{n}r_i(t)$.  The total cumulative reward up to time $t$ is
$CR(t)= \sum_{\tau=1}^{t}R(\tau)$.  During time step $t$, the RNNAI
produces an output action vector, $out(t) \in \mathbb{R}^o$, which may
influence the environment and thus future $sense(\tau)$ for $\tau >t$.
At any given time, the RNNAI's goal is to maximize $CR(t_{death})$.

Let $all(t) \in \mathbb{R}^{m+n+o}$ denote the concatenation of
$sense(t)$ and $out(t)$.  Let $H(t)$ denote the sequence $(all(1),
all(2), \ldots, all(t))$ up to time $t$.

To be able to retrain its components on all observations ever made,
 {\em
  the RNNAI stores its entire, growing, lifelong sensory-motor interaction
  history $H(\cdot)$ including all inputs and actions and reward signals observed
  during all successful and failed trials~\cite{Schmidhuber:06cs,Schmidhuber:09sice},
  including what initially looks like noise but later may turn out to
  be regular}. This is normally not done, but is feasible today. 
  
  That is, all data is ``holy'', and never discarded, in line with
  what mathematically optimal general problem solvers should do
  \cite{Hutter:05book+,Schmidhuber:02colt}.
  Remarkably, even human brains may have enough storage capacity
to store 100 years of sensory input at a reasonable resolution~\cite{Schmidhuber:09sice}.

\subsection{Standard Activation Spreading in Typical RNNs}
\label{rnn}

Many RNN-like models can be used to build general computers, e.g.,
neural pushdown automata~\cite{Das:92,mozer1993connectionist}, NNs
with quickly modifiable, differentiable external memory based on fast
weights~\cite{Schmidhuber:92ncfastweights}, or closely related
RNN-based meta-learners~\cite{Schmidhuber:93selfreficann,Schmidhuber:93ratioicann,Hochreiter:01meta,scholarpedia2010}.
Using sloppy but convenient terminology, we refer to all of them as
RNNs.  A typical implementation of $M$ uses an LSTM network (see
Sec.~\ref{SLUL}).  If there are large 2-dimensional inputs such as
video images, then they can be first filtered through a CNN (compare
Sec.~\ref{SLUL} and \ref{pre}) before fed into the LSTM.  Such a
CNN-LSTM combination is still an RNN.

Here we briefly summarize information processing in standard RNNs.
Using notation similar to the one of a previous
survey~\cite[Sec.~2]{888}, let $i,k,s$ denote positive integer
variables assuming ranges implicit in the given contexts.  Let
$n_u,n_w,T$ also denote positive integers.

At any given moment, an RNN (such as the $M$ of Sec.~\ref{M}) can be
described as a connected graph with $n_u$ units (or nodes or neurons)
in a set $N=\{u_1,u_2, \ldots,u_{n_u}\}$ and a set $H \subseteq N
\times N$ of directed edges or connections between nodes.  The input
layer is the set of input units, a subset of $N$.  In fully connected
RNNs, all units have connections to all non-input units.

The RNN's behavior or program is determined by $n_w$ real-valued,
possibly modifiable, parameters or weights, $w_i$ $(i=1,\ldots,n_w)$.
During an episode of information processing (e.g., during a trial of
Sec.~\ref{scheme}), there is a {\em partially causal sequence} $x_s
(s=1,\ldots,T)$ of real values called events.  Here the index $s$ is
used in a way that is much more fine-grained than the one of the index
$t$ in Sec.~\ref{holy}, \ref{M}, \ref{CM}: a single time step may
involve numerous events.  Each $x_s$ is either an input set by the
environment, or the activation of a unit that may directly depend on
other $x_k (k<s)$ through a current NN topology-dependent set, $in_s$,
of indices $k$ representing incoming causal connections or links.  Let
the function $v$ encode topology information, and map such event index
pairs, $(k,s)$, to weight indices.  For example, in the non-input case
we may have $x_s=f_s(net_s)$ with real-valued $net_s=\sum_{k \in in_s}
x_k w_{v(k,s)}$ (additive case) or $net_s=\prod_{k \in in_s} x_k
w_{v(k,s)}$ (multiplicative case), where $f_s$ is a typically
nonlinear real-valued {\em activation function} such as $tanh$.  Other
$net$ functions combine additions and
multiplications~\cite{ivakhnenko1965,ivakhnenko1971}; many other
activation functions are possible.  The sequence, $x_s$, may directly
affect certain $x_k (k>s)$ through outgoing connections or links
represented through a current set, $out_s$, of indices $k$ with $s \in
in_k$.  Some of the non-input events are called {\em output events}.

Many of the $x_s$ may refer to {\em different}, time-varying
activations of the {\em same} unit, e.g., in RNNs.  During the
episode, the same weight may get reused over and over again in
topology-dependent ways.  Such weight sharing {\em across space and/or
  time} may greatly reduce the NN's descriptive complexity, which is
the number of bits of information required to describe the NN
(Sec.~\ref{M}).  Training algorithms for the RNNs of our RNNAIs will
be discussed later.

\subsection{Alternating Training Phases for  Controller $\boldsymbol{C}$ and  World Model $\boldsymbol{M}$}
\label{scheme}

 Several novel implementations of $C$ are
described in Sec.~\ref{CM}.  All of them make use of a variable size
RNN called the world model, $M$, which learns to compactly encode the growing
history, for example, through predictive coding, trying to predict (the expected
value of) each input component, given the history of actions and
observations.  $M$'s goal is to discover algorithmic regularities in
the data so far by learning a program that compresses the data better
in a lossless manner.  Example details will be specified in Sec.~\ref{M}.

Both $C$ and $M$ have real-valued parameters or weights that can be
modified to improve performance.  To avoid instabilities, $C$ and $M$
are trained in alternating fashion, as in Algorithm~\ref{alternate}.

\begin{algorithm}[t]
\begin{algorithmic}
\STATE 1. Initialize  $C$ and $M$ and their weights.
\STATE 2. Freeze $M$'s weights such that they cannot change while $C$ learns.
\STATE 3. Execute a new trial by generating a finite action sequence that prolongs the history of 
actions and observations. 
Actions may be due to $C$ which may exploit $M$ in various ways (see Sec.~\ref{CM}). 
Train $C$'s weights  on the prolonged (and recorded) history to generate action sequences
with higher expected reward, using methods of Sec.~\ref{CM}.
\STATE 4. Unfreeze $M$'s weights, and re-train $M$ in a ``sleep phase" to better predict/compress the prolonged history; see Sec.~\ref{M}.
\STATE 5. If no stopping criterion is met, goto 2.
\end{algorithmic}
\caption{Train $C$ and $M$ in Alternating Fashion}
\label{alternate}
\end{algorithm}

\section{The Gradient-Based World Model $\boldsymbol{M}$}
\label{M}

A central objective of unsupervised learning is to compress the observed data~\cite{Barlow:89,chunker91and92}.
$M$'s goal is to compress the RL agent's entire growing
interaction history of all failed and successful trials~\cite{Schmidhuber:06cs,Schmidhuber:10ieeetamd}, e.g., through predictive coding~\cite{chunker91and92,SchmidhuberHeil:96}.
$M$ has $m+n+o$ input units to receive $all(t)$ at time $t <
t_{death}$, and $m+n$ output units to produce a prediction $pred(t+1)
\in \mathbb{R}^{m+n}$ of $sense(t+1)$
~\cite{Schmidhuber:90sandiego,Schmidhuber:90sab,Schmidhuber:90cmss,Schmidhuber:91nips}.

\subsection{$\boldsymbol{M}$'s Compression Performance on the History so far}
\label{compression}

Let us address details of training $M$ in a ``sleep phase" of step 4 in
algorithm~\ref{alternate}.  (The training of $C$ will be discussed in
Sec.~\ref{CM}.)  Consider some $M$ with given (typically suboptimal)
weights and a default initialization of all unit activations.  
One example way of making $M$ compress the history (but not the only one) is the following.
Given
$H(t)$, we can train $M$ by replaying~\cite{Lin:91} $H(t)$ in
semi-offline training, sequentially feeding $all(1), all(2), \ldots
all(t)$ into $M$'s input units in standard RNN fashion (Sec.
\ref{SLUL},~\ref{rnn}).  Given $H(\tau)$ ($\tau<t$), $M$ calculates $pred(\tau +
1)$, a prediction of $sense(\tau+1)$.  A standard error function to be
minimized by gradient descent in $M$'s weights (Sec.~\ref{SLUL})
would be $E(t) = \sum_{\tau=1}^{t-1} \| pred(\tau+1) - sense(\tau +1)
\|^2$, the sum of the deviations of the predictions from the
observations so far.

However, $M$'s goal is not only to minimize the total prediction error,
$E$.  Instead, to avoid the erroneous ``discovery" of ``regular
patterns" in irregular noise, we use AIT's sound way of dealing with
overfitting~\cite{Solomonoff:64,Kolmogorov:65,Wallace:68,Rissanen:86,LiVitanyi:97,gruenwald2005},
and measure $M$'s compression performance by the number of bits
required to specify $M$, plus the bits needed to encode the observed
deviations from $M$'s
predictions~\cite{Schmidhuber:06cs,Schmidhuber:10ieeetamd}.  For
example, whenever $M$ incorrectly predicts certain input pixels of a
perceived video frame, those pixel values will have to be encoded
separately, which will cost storage space.  (In typical applications,
$M$ can only execute a fixed number of elementary computations per
time step to compress and decompress data, which usually has to be
done online. That is, in general $M$ will not reflect the data's true
Kolmogorov complexity~\cite{Solomonoff:64,Kolmogorov:65}, but at best
a time-bounded variant thereof~\cite{LiVitanyi:97}.)

Let integer variables, $bits_M$ and  $bits_H$, denote estimates of the number
of bits required to encode (by a fixed algorithmic scheme) the current
$M$, and the deviations of $M$'s predictions from the observations on
the current history, respectively.  For example, to obtain $bits_H$,
we may naively assume some simple, bell-shaped, zero-centered
probability distribution $P_e$ on the finite number
of possible real-valued prediction errors $e_{i,\tau}= (pred_i(\tau) -
sense_i(\tau))^2$ (in practical applications the errors will be given
with limited precision), and encode each $e_{i,\tau}$ by $-log
P_e(e_{i,\tau})$ bits~\cite{Huffman:52,Shannon:48}.  That is, large
errors are considered unlikely and cost more bits than small ones.  To
obtain $bits_M$, we may naively multiply the current number of $M$'s
non-zero modifiable weights by a small integer constant reflecting the
weight precision.  Alternatively, we may assume some simple,
bell-shaped, zero-centered probability distribution, $P_w$, on the
finite number of possible weight values (given with limited
precision), and encode each $w_i$ by $-log P_w(w_i)$ bits.  That is,
large absolute weight values are considered unlikely and cost more
bits than small ones~\cite{Hanson:89,Weigend:91,Krogh:92,Hinton:93}.
Both alternatives ignore the possibility that $M$'s entire weight
matrix might be computable by a short computer
program~\cite{Schmidhuber:97nn+,koutnik:gecco13}, but have the
advantage of being easy to calculate.  Moreover, since $M$ is a
general computer itself, at least in principle it has a chance of
learning equivalents of such short programs.

\subsection{$\boldsymbol{M}$'s Training}
\label{trainM}

To decrease $bits_M + bits_H$, we add a {\em regularizing} term to
$E$, to punish excessive
complexity~\cite{Akaike:70,akaike1974,Hanson:89,Weigend:91,MacKay:92b,Krogh:92,Hinton:93,Moody:94a,Moody:92,Holden:94,Wang:94,Amari:93,Wang:94,Vapnik:92a,Vapnik:92,Wolpert:94b,Hochreiter:97nc1,Hochreiter:99nc}.

Step 1 of algorithm~\ref{alternate} starts with a small $M$.  As the
history grows, to find an $M$ with small $bits_M + bits_H$, step
4 uses {\em sequential network construction}: it regularly changes
$M$'s size by adding or pruning units and
connections~\cite{ivakhnenko1968,ivakhnenko1971,Ash:89,Moody:89,gallant1988,honavar1988,White:89,Mozer:89a,LeCun:90a,Hassibi:93,Levin:94,Ring:91,Fahlman:91,weng1992,honavar1993,burgess1994,fritzke94,parekh2000,utgoff2002}.
Whenever this helps (after additional training with BPTT of $M$---see
Sec.~\ref{SLUL}) to improve $bits_M + bits_H$ on the history so far,
the changes are kept, otherwise they are discarded.  (Note that even animal
brains grow and prune neurons.)

Given history $H(t)$, instead of re-training $M$ in a sleep phase
(step 4 of algorithm~\ref{alternate}) on all of $H(t)$, we may
re-train it on parts thereof, by selecting trials randomly or
otherwise from $H(t)$, and replay them to retrain $M$ in standard
fashion (Sec.~\ref{SLUL}).  To do this, however, all of $M$'s unit
activations need to be stored at the beginning of each trial.  ($M$'s
hidden unit activations, however, do not have to be stored if they are
reset to zero at the beginning of each trial.)

\subsection{$\boldsymbol{M}$ may have a Built-In FNN Preprocessor}
\label{pre}

To facilitate $M$'s task in certain environments, each frame of the
sensory input stream (video, etc.) can first be separately compressed
through autoencoders~\cite{Rumelhart:86} or autoencoder
hierarchies~\cite{ballard1987modular,bengio2013tpami} based on CNNs or
other FNNs (see Sec.~\ref{SLUL}) \cite{ciresan2012cvpr} used as
sensory preprocessors to create less redundant sensory
codes~\cite{Jodogne07,lange,Legenstein2010,cuccu2011}.  The compressed
codes are then fed into an RNN trained to predict not the raw inputs,
but their compressed codes.  Those predictions have to be decompressed
again by the FNN, to evaluate the total compression performance,
$bits_M + bits_H$, of the FNN-RNN combination representing $M$.

\begin{figure}[ht]
\begin{center}
\includegraphics[width=\linewidth]{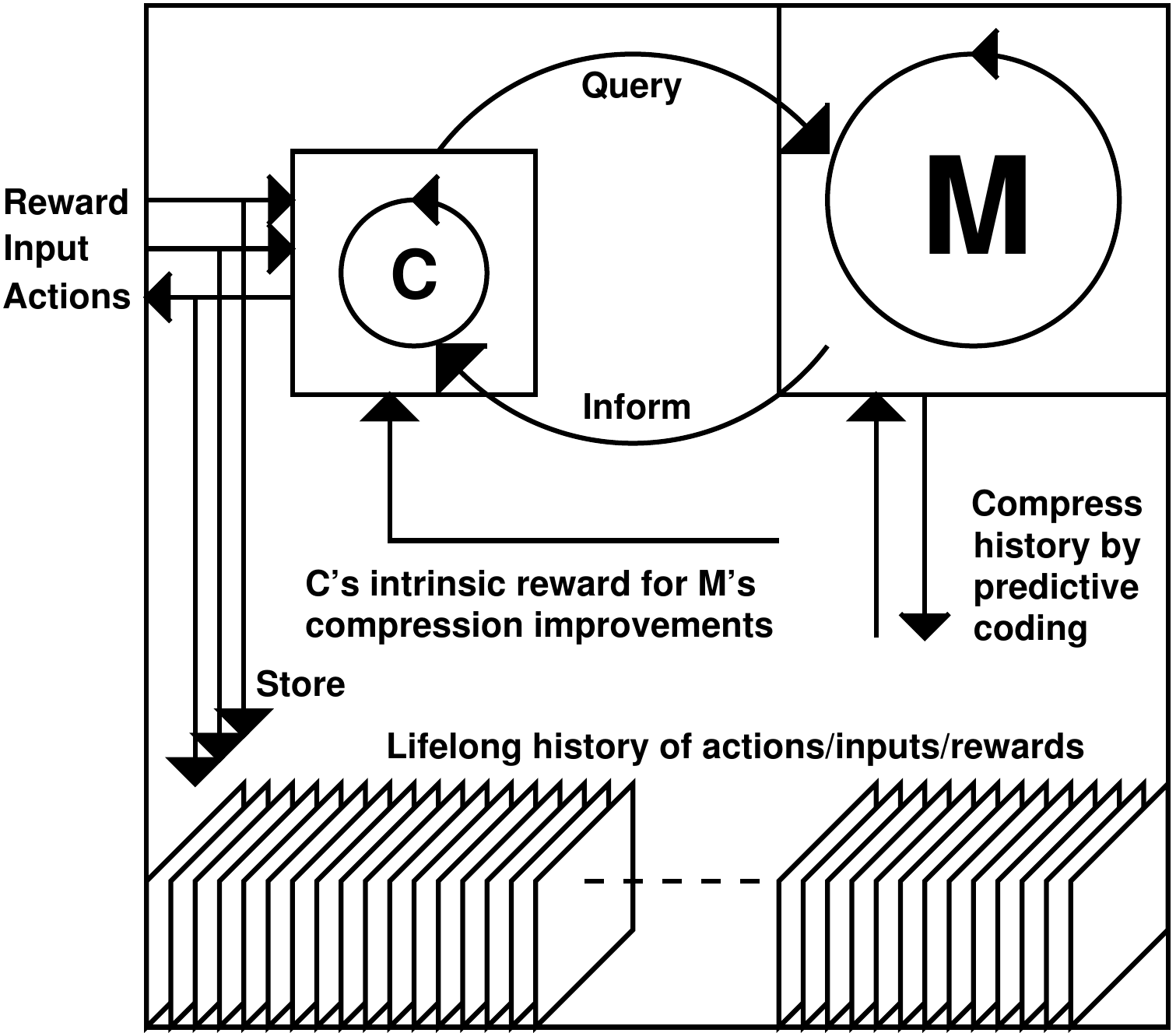}
\end{center}
\caption{{\em In a series of trials, an RNN controller $C$ steers an
    agent interacting with an initially unknown, partially observable environment.  The
    entire lifelong interaction history is stored, and used to train
    an RNN world model $M$, which learns to predict new inputs from
    histories of previous inputs and actions, using predictive coding
    to compress the history (Sec.~\ref{M}). Given an RL problem, $C$
    may speed up its search for rewarding behavior by learning
    programs that address/query/exploit $M$'s program-encoded
    knowledge about predictable regularities, e.g., through 
extra connections from and to (a copy of) $M$---see Sec.~\ref{think}.
This may be much cheaper than learning reward-generating programs from
    scratch.  $C$ also may get intrinsic reward for creating
    experiments causing data with yet unknown regularities that
    improve $M$ (Sec.~\ref{curiosity}).  Not shown are deep FNNs as
    preprocessors (Sec.~\ref{pre}) for high-dimensional data (video
    etc) observed by $C$ and $M$.  }}
\label{fig}
\end{figure}

\section{The Controller $\boldsymbol{C}$ Learning to Exploit RNN World Model $\boldsymbol{M}$}
\label{CM}

Here we describe ways of using the world model, $M$, of Sec.~\ref{M} to
facilitate the task of the RL controller, $C$.  Especially the systems
of Sec.~\ref{think} overcome drawbacks of early $CM$ systems mentioned
in Sec.~\ref{earlyCM},~\ref{earlyMtradRL}.
Some of the setups of the present Sec. \ref{CM} can be 
viewed as  special cases of the general scheme in Sec. \ref{general}.

\subsection{$\boldsymbol{C}$ as a Standard RL Machine whose States are $\boldsymbol{M}$'s Activations}
\label{standard}

We start with details of an approach whose principles date back to the
early
1990s~\cite{Schmidhuber:91nips,Lin92memoryapproaches} (Sec.~\ref{earlyMtradRL}).
Given an RNN or RNN-like $M$ as in Sec.~\ref{M}, we implement $C$ as a
traditional RL machine~\cite{Sutton:98}\cite[Sec.~6.2]{888}
based on the Markov assumption (Sec.~\ref{GDRL}).  While $M$ is
processing the history of actions and observations to predict future
inputs, the internal states of $M$ are used as inputs to a predictor
of cumulative expected future reward.

More specifically, in step 3 of algorithm~\ref{alternate},
consider a trial lasting from time $t_a \geq 1$ to $t_b \leq
t_{death}$.  $M$ is used as a preprocessor for $C$ as follows.  At the
beginning of a given time step, $t$, of the trial $(t_a\leq t<t_b)$, let
$hidden(t) \in \mathbb{R}^{h}$ denote the vector of
$M$'s  current hidden unit activations (those units that are neither input
nor output units).  Let $state(t) \in \mathbb{R}^{2m+2n+h}$ denote the
concatenation of $sense(t)$, $hidden(t)$ and $pred(t)$.  (In cases where
$M$'s activations are reset after each trial, $hidden(t_a)$ and
$pred(t_a)$ are initialized by default values, e.g., zero vectors.)

$C$ is an RL machine with $2m+2n+h$-dimensional inputs and
$o$-dimensional outputs.  At time $t$, $state(t)$ is fed into $C$,
which then computes action $out(t)$.  Then $M$ computes from
$sense(t)$, $hidden(t)$ and $out(t)$ the values $hidden(t+1)$ and
$pred(t+1)$.  Then $out(t)$ is executed in the environment, to obtain
the next input $sense(t+1)$.

The parameters or weights of $C$ are trained to maximize reward by a
standard RL method such as Q-learning or similar
methods~\cite{BartoSuttonAnderson:83,Watkins:89,WatkinsDayan:92,Moore:93,Schwartz:93,Rummery:94,Singh:94R,Baird:95,Kaelbling:95,Peng:96,Mahadevan:96,Tsitsiklis:96,96-BradtkeLstd,Santamaria:97,prwu97,Sutton:98,Wiering:98,baird:nips12,meuleau:icuai99,Doya:00,Bertsekas:01,brafman02,Abounadi:02,03-LspiLagoudakis,09-Gtd,10-GqLambda,hasselt2012}.
Note that most of these methods evaluate not only input events but 
pairs of input and output
(action) events.

In one of the simplest cases, $C$ is just a linear perceptron FNN
(instead of an RNN like in the early
system~\cite{Schmidhuber:91nips}).  The fact that $C$ has no built-in
memory in this case is not a fundamental restriction since $M$ is
recurrent, and has been trained to predict not only normal sensory
inputs, but also reward signals.  That is, the state of $M$ must
contain all the historic information relevant to maximize future
expected reward, provided the data history so far already contains
the relevant experience, and $M$ has learned to compactly
extract and represent its regular aspects.
 
This approach is different from other, previous combinations of traditional
RL~\cite{Sutton:98}\cite[Sec.~6.2]{888} and
RNNs~\cite{Schmidhuber:91nips,Lin:93,Bakker:03robot} which use RNNs
only as value function approximators that directly predict cumulative
expected reward, instead of trying to predict all sensations time step
by time step.  The $CM$ system in the present section 
separates the hard task of prediction in partially observable environments from the comparatively
simple task of RL under the Markovian assumption that the current
input to $C$ (which is $M$'s state) contains all information relevant
for achieving the goal.
 
 \subsection{$\boldsymbol{C}$ as an Evolutionary RL (R)NN whose Inputs are $\boldsymbol{M}$'s Activations}
\label{evolution}

This approach is essentially the same as the one of Sec.
\ref{standard}, except that $C$ is now an FNN or RNN trained by
evolutionary algorithms
~\cite{Rechenberg:71,Schwefel:74,Holland:75,Fogel:66,goldberg:gabook89}
applied to NNs
~\cite{miller:icga89,yao:review93,nolfi:alife4,Sims:1994:EVC,Gomez:08jmlr,hansenCMA,hansen2003,igel:cec03,heidrich-meisner:09},
or by policy gradient methods
~\cite{Williams:86,Williams:88,Williams:92,Sutton:99,baxter2001,aberdeenthesis,ghavamzadehICML03,stoneICRA04,wierstraCEC08,rueckstiess2008b,peters2008neuralnetworks,peters2008neurocomputing,sehnke2009parameter,gruettner2010multi,wierstra2010,peters2010,grondman2012,heess2012}\cite[Sec.~6.6]{888},
or by Compressed NN Search; see Sec.~\ref{intro}.  $C$ has $2m+2n+h$
input units and $o$ output units.  At time $t$, $state(t)$ is fed into
$C$, which computes $out(t)$; then $M$ computes $hidden(t+1)$ and
$pred(t+1)$; then $out(t)$ is executed to obtain $sense(t+1)$.

\subsection{$\boldsymbol{C}$ Learns to Think with $\boldsymbol{M}$: High-Level Plans and Abstractions}
\label{think}

Our RNN-based $CM$ systems of the early
1990s~\cite{Schmidhuber:90sandiego,Schmidhuber:90sab}(Sec.~\ref{earlyCM})
could in principle plan ahead by performing numerous fast {\em mental}
experiments on a predictive RNN world model, $M$, instead of
time-consuming real experiments, extending earlier work on reactive
systems without memory~\cite{Werbos:89identification,Sutton:90dyna}.   
However, this can work well only in
(near-)deterministic environments, and, even there, $M$ would have to
simulate many entire alternative futures, time step by time step, to
find an action sequence for $C$ that maximizes reward.  This method
seems very different from the much smarter hierarchical planning
methods of humans, who apparently can learn to identify and exploit a
few relevant problem-specific abstractions of possible future events;
reasoning abstractly, and efficiently ignoring irrelevant
spatio-temporal details.

We now describe a $CM$ system that can in principle learn to plan and
reason like this as well, according to the AIT argument
(Sec.~\ref{argument}).  This should be viewed as a main contribution
of the present paper.  See Figure \ref{fig}.

Consider an RNN $C$ (with typically rather
small feasible search space) as in Sec.~\ref{evolution}.  We 
add standard and/or multiplicative
learnable connections (Sec.~\ref{rnn}) from some of the units of $C$
to some of the units of the typically huge unsupervised $M$, and from some of the units of $M$ to
some of the units of $C$.  The new connections are said to belong to
$C$.  $C$ and $M$ now collectively form a new RNN called $CM$, with
standard activation spreading as in Sec.~\ref{rnn}.  The activations
of $M$ are initialized to default values at the beginning of each
trial.
Now $CM$ is trained on RL tasks in line with
step 3 of algorithm~\ref{alternate}, using search methods such as
those of Sec.~\ref{evolution} (compare Sec.~\ref{intro}).  The
(typically many) connections of $M$, however, do not change---only
the (typically relatively few) connections of $C$ do.

What does that mean? It means that now $C$'s relatively small
candidate programs are given time to ``think'' by feeding sequences of
activations into $M$, and reading activations out of $M$, before and
while interacting with the environment.  Since $C$ and $M$ are
general computers, $C$'s programs may query, edit or invoke
subprograms of $M$ in arbitrary, computable ways through
the new connections. Given some RL problem, according to the AIT
argument (Sec.~\ref{argument}), this can greatly accelerate $C$'s
search for a problem-solving weight vector $\hat{w}$, provided the
(time-bounded~\cite{LiVitanyi:97}) mutual algorithmic information
between $\hat{w}$ and $M$'s program is high, as is to be expected in
many cases since $M$'s environment-modeling program should reflect
many regularities useful not only for prediction and coding, but also
for decision making.\footnote{ An alternative way of letting $C$ learn
  to access the program of $M$ is to add $C$-owned connections from
  the {\em weights} of $M$ to units of $C$, treating the current
  weights of $M$ as additional real-valued inputs to $C$.  This,
  however, will typically result in a much larger search space for
  $C$. There are many other
variants of the general scheme described in Sec. \ref{general}.
  }

This simple but novel approach is much more general than previous
computable, but restricted, ways of letting a feedforward $C$ use a
model $M$
(Sec.~\ref{earlyCM})\cite{Werbos:89identification,Sutton:90dyna}\cite[Sec.~6.1]{888},
by simulating entire possible futures step by step, then propagating
error signals or temporal difference errors backwards (see Section
\ref{earlyCM}).  Instead, we give $C$'s program search an opportunity
to discover sophisticated computable ways of exploiting $M$'s code,
such as abstract hierarchical planning and analogy-based reasoning.  For
example, to represent previous observations, an $M$ implemented as an
LSTM network (Sec.~\ref{SLUL}) will develop high-level, abstract,
spatio-temporal feature detectors that may be active for thousands of
time steps, as long as those memories are useful to predict (and thus compress) future
observations~\cite{Gers:02jmlr,Gers:2000nc,Perez:02,graves:2009nips}.
However, $C$ may learn to directly invoke the corresponding
``abstract'' units in $M$ by inserting appropriate pattern sequences
into $M$.  $C$ might then short-cut from there to typical {\em
  subsequent} abstract representations, ignoring the long input
sequences normally required to invoke them in $M$, thus quickly
anticipating a few possible positive outcomes to be pursued (plus
computable ways of achieving them), or negative outcomes to be
avoided.

Note that $M$ (and by extension $M$) does not at all have to be a
perfect predictor.  For example, it won't be able to predict noise.
Instead $M$ will have learned to approximate conditional expectations of future inputs,
given the history so far. A naive way of 
exploiting  $M$'s probabilistic
knowledge would be to plan ahead through naive step-by-step Monte-Carlo simulations
of possible $M$-predicted futures, to find and execute action sequences that maximize expected reward predicted by
those simulations. However, we won't limit the system to this naive approach. 
Instead
it will be the task of $C$ to learn to address useful problem-specific
parts of the current $M$, and reuse them for problem solving.
Sure, $C$ will have to intelligently exploit $M$, which will cost
bits of information (and thus search time for appropriate weight
changes of $C$), but this is often still much cheaper in the AIT sense
than learning a good $C$ program from scratch, as in our previous
non-RNN AIT-based work on {\em algorithmic transfer learning}~\cite{Schmidhuber:04oops}, 
where self-invented
recursive code for previous solutions sped up the search for code for
more complex tasks by a factor of 1000.

Numerous topologies are possible for the adaptive connections from $C$
to $M$, and back.  Although in some applications $C$ may find it hard
to exploit $M$, and might prefer to ignore $M$ (by setting
connections to and from $M$ to zero), in some environments under
certain $CM$ topologies, $C$ can greatly profit from $M$.

While $M$'s weights
are frozen in step 3 of algorithm~\ref{alternate}, the weights of $C$ can
learn when to make $C$ attend to history information represented by
$M$'s state, and when to ignore such information, and instead use $M$'s
innards in other computable ways.  This can be further facilitated by
introducing a special unit, $\hat{u}$, to $C$, where $\hat{u}(t)all(t)$
instead of $all(t)$ is fed into $M$ at time $t$, such that $C$ can
easily (by setting $\hat{u}(t)=0$) force $M$ to completely ignore
environmental inputs, to use $M$ for ``thinking'' in other ways.

Should $M$ later grow (or shrink) in step 4 of algorithm~\ref{alternate}, 
in line with Sec.~\ref{trainM},
$C$ may in turn grow additional connections to and from $M$ (or lose some)
in the next incarnation of step 3.

\subsection{Incremental / Hierarchical / Multitask Learning of $\boldsymbol{C}$ with $\boldsymbol{M}$}
\label{incremental}

A variant of the approach in Sec.~\ref{think} incrementally trains $C$
on a never-ending series of tasks, continually building on solutions
to previous problems, instead of learning each new problem from
scratch.  In principle, this can be done
through incremental NN evolution~\cite{gomez:ab97},
hierarchical NN evolution~\cite{stoneML05,vanhoorn:09cig},
hierarchical Policy Gradient
algorithms~\cite{ghavamzadehICML03},
or  asymptotically optimal ways of {\em algorithmic transfer learning}~\cite{Schmidhuber:04oops}.

Given a new task and a $C$ trained on several previous tasks, such
hierarchical/incremental methods may freeze the current weights of
$C$, then enlarge $C$ by adding new units and connections which are
trained on the new task.  This process reduces the size of the search
space for the new task,  giving the new weights the opportunity to
learn to use the frozen parts of $C$ as subprograms.

Incremental variants of {\em Compressed RNN
  Search}~\cite{koutnik:gecco13} (Sec.~\ref{intro}) do not directly
search in $C$'s potentially large weight space, but in the frequency
domain by representing the weight matrix as a small set of Fourier-type
coefficients. By searching for new coefficients to be added to already
learned set responsible for solving previous problems, $C$'s weight
matrix is fine tuned incrementally and indirectly (through
superpositions).  Given a current problem, in AIT-based OOPS
style~\cite{Schmidhuber:04oops}, we may impose growing run time limits
on programs tested on $C$, until a solution is found.

\section{Exploration: Rewarding $\boldsymbol{C}$ for Experiments that Improve  $\boldsymbol{M}$}  
\label{curiosity}
Humans, even as infants, invent their own tasks in a curious and
creative fashion, continually increasing their problem solving
repertoire even without an external reward or teacher.  They seem to
get intrinsic reward for creating experiments leading to observations
that obey a previously unknown law that allows for better compression
of the observations---corresponding to the discovery of a {\em
  temporarily interesting, subjectively novel
  regularity}~\cite{Schmidhuber:91singaporecur,Schmidhuber:06cs,Schmidhuber:10ieeetamd}
(compare also~\cite{Singh:05nips,Oudeyer:12intrinsic}).

For example, a video of 100 falling apples can be greatly compressed
via predictive coding once the law of gravity is discovered.
Likewise, the video-like image sequence perceived while moving through
an office can be greatly compressed by constructing an internal 3D
model of the office space~\cite{Schmidhuber:13powerplay}. The 3D model
allows for re-computing the entire high-resolution video from a
compact sequence of very low-dimensional eye coordinates and eye
directions. The model itself can be specified by far fewer
bits of information than needed to store the raw pixel data of a long
video. Even if the 3D model is not precise, only relatively few
extra bits will be required to encode the observed deviations from the
predictions of the model.

Even mirror
  neurons~\cite{kohler2002} are easily explained as by-products of
  history compression as in Sec.~\ref{holy} and \ref{M}.  They fire
  both when an animal acts and when the animal observes the same
  action performed by another. Due to mutual algorithmic information
  shared by perceptions of similar actions performed by various
  animals, efficient RNN-based predictive coding (Sec.~\ref{holy},
  \ref{M}) profits from using the same feature detectors (neurons) to
  encode the shared information, thus saving storage space. 

Given the $C$-$M$ combinations of Sec.~\ref{CM}, we motivate $C$ to
become an efficient explorer and an {\em artificial scientist}, by
adding to its standard {\em external} reward (or fitness) for solving
user-given tasks another {\em intrinsic} reward for generating novel
action sequences ($=$ experiments) that allow $M$ to improve its
compression performance on the resulting
data~\cite{Schmidhuber:06cs,Schmidhuber:10ieeetamd}.

At first glance, repeatedly evaluating $M$'s compression performance
on the entire history seems impractical.  A heuristic to overcome this
is to focus on $M$'s improvements on the most recent trial, while
regularly re-training $M$ on randomly selected previous trials, to
avoid catastrophic forgetting.

A related problem is that $C$'s incremental program search may find it
difficult to identify (and assign credit to) those parts of $C$
responsible for improvements of a huge, black box-like, monolithic
$M$.  But we can implement $M$ as a self-modularizing, computation
cost-minimizing, winner-take-all
RNN~\cite{Schmidhuber:89cs,Schmidhuber:12slimnn,Srivastava2013first}.
Then it is possible to keep track of which parts of $M$ are used to
encode which parts of the history.  That is, to evaluate weight
changes of $M$, only the affected parts of the stored history have to
be re-tested~\cite{Schmidhuber:13powerplay}.  Then $C$'s search can be
facilitated by tracking which parts of $C$ affected those parts of
$M$.  By penalizing $C$'s programs for the time consumed by such
tests, the search for $C$ is biased to prefer programs that conduct
experiments causing data yielding {\em quickly verifiable} compression
progress of $M$.  That is, the program search will prefer to change
weights of $M$ that are {\em not} used to compress large parts of the
history that are expensive to
verify~\cite{Schmidhuber:12slimnn,Schmidhuber:13powerplay}.  The first
implementations of this simple principle were described in our work on
the {\sc PowerPlay}
framework~\cite{Schmidhuber:13powerplay,Srivastava2013first}, which
incrementally {\em searches the space of possible pairs of new tasks
  and modifications of the current program,} until it finds a more
powerful program that, unlike the unmodified program, solves all
previously learned tasks plus the new one, or
simplifies/compresses/speeds up previous solutions, without forgetting
any.  Under certain conditions this can accelerate the acquisition of
external reward specified by user-defined tasks.

\section{Conclusion}
We introduced novel combinations of a reinforcement learning (RL) controller, $C$, and an
RNN-based predictive world model, $M$.  The most general $CM$ systems
implement principles of  {\em algorithmic}~\cite{Solomonoff:64,Kolmogorov:65,LiVitanyi:97}
as opposed to {\em traditional}~\cite{boltzmann1909,Shannon:48}
information theory. Here both $M$ and $C$ are RNNs or RNN-like systems.  $M$ is actively exploited
in arbitrary computable ways by $C$, whose program search space is
typically much smaller, and which may learn to selectively probe and reuse $M$'s internal
programs to plan and reason.  
The basic principles are not limited to RL, but apply to
all kinds of active algorithmic transfer learning from one RNN to another.
By combining gradient-based RNNs and
RL RNNs, we create a qualitatively new type of self-improving, general
purpose, connectionist control architecture.  This RNNAI may continually build
upon previously acquired problem solving procedures, some of them
self-invented in a way that resembles a scientist's search for novel
data with unknown regularities, preferring still-unsolved but quickly
learnable tasks over others.


\newpage
\bibliography{bib}
\bibliographystyle{abbrv}
\end{document}